\documentclass[runningheads]{llncs}

 \usepackage{eccv}

\usepackage{eccvabbrv}
\usepackage{adjustbox}
\usepackage{graphicx}
\usepackage{booktabs}
\usepackage{multirow}
\usepackage{subcaption}
\usepackage{makecell}
\usepackage{tabularx}
\usepackage{array}
\usepackage{siunitx}

\usepackage{pifont}

\usepackage[accsupp]{axessibility}

\usepackage{placeins}

\usepackage{xcolor}

\definecolor{darkblue}{rgb}{0.21,0.49,0.74}

\usepackage{hyperref}

\hypersetup{
    colorlinks=true,
    linkcolor=black,
    citecolor=darkblue,
    filecolor=black,
    urlcolor=darkblue
}

\usepackage{orcidlink}

\begin{document}

\title{Patch-Derived or Register Tokens? Comparing Variable-Length Image Tokenizers}
\titlerunning{Comparing Variable-Length Image Tokenizers}

\author{
    Philipp Schenkel\inst{1}\orcidlink{0009-0006-6507-8599}
    \and
    Tobias Sch\"{u}rmann\inst{1}\orcidlink{0000-0002-1967-7076}
    \and
    \"{O}mer \c{S}ahin Ta\c{s}\inst{1,2}\orcidlink{0000-0002-1249-260X}
}

\authorrunning{P. Schenkel et al.}

\institute{
    FZI Research Center for Information Technology\\
    \and
    Karlsruhe Institute of Technology\\
}

\maketitle

\vspace{-1.7em}
\begin{figure}[htbp]
    \centering
    \includegraphics[width=\textwidth,    height=0.25\textheight,
        keepaspectratio]{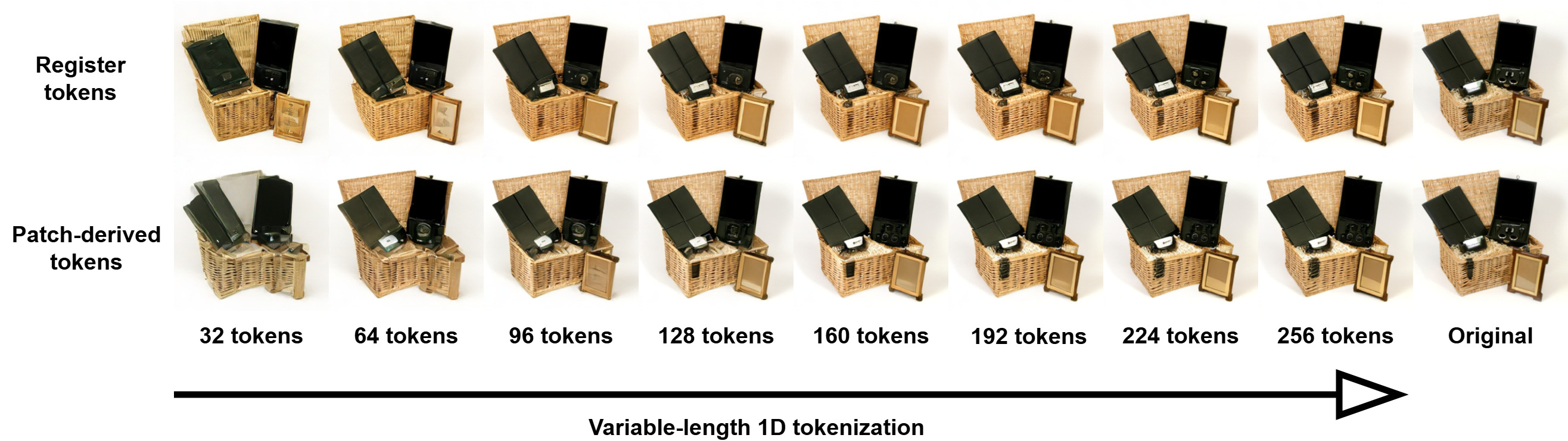}
    \caption{Qualitative reconstructions from two trained image tokenizers: one using patch-derived tokens and the other using register tokens. Both models progressively recover finer visual details as more tokens become available, while organizing information differently across the latent sequence.}
    \vspace{-3.0em}
\end{figure}

\begin{abstract}
Image tokenizers convert pixels into compact latent sequences for efficient generative modeling. Most approaches arrange latents on a downsampled two-dimensional grid, tying sequence length to image resolution. 
We compare two variable-length strategies: computationally efficient patch-derived tokens, which inherit explicit spatial organization, and more demanding register tokens, which have no predefined spatial layout. 
Our reconstruction results reveal that patch-derived tokens perform substantially better under extreme compression, whereas register tokens achieve slightly better reconstruction quality at intermediate budgets. At high token budgets, both approaches achieve similar reconstruction quality. For downstream generative tasks, register tokens outperform patch-derived tokens. 
Adding representation alignment with external pretrained visual encoders substantially improves the generative performance of patch-derived tokens, allowing them to match or outperform register tokens depending on the token budget. 
We further find that both approaches learn spatially localized representations. 
Patch-derived tokens largely retain their explicitly imposed spatial organization, while register tokens develop an implicit two-dimensional spatial structure despite having no predefined layout. 
Representation alignment reduces this spatial organizational constraint, enabling individual tokens to exert a stronger non-local influence.

\end{abstract}

\section{Introduction}

Image tokenization plays a crucial role in reducing the computational costs of subsequent generative models \cite{sun2024, chang2022, rombach2022}. An image tokenizer \cite{oord2017} encodes an image into a compact latent token representation. Most tokenizers preserve the spatial structure of the image by arranging the tokens on a downsampled grid, with each token corresponding to a fixed image region \cite{esser2021, mentzer2024}. Consequently, the number of tokens grows with the image resolution, resulting in prohibitively long sequences in the case of high-resolution image generation, video generation, and world modeling.

Recent work such as TiTok \cite{yu2024} has challenged these grid-structured latent representations by encoding images into sequences whose length is not determined by the number of input patches. This decoupling enables substantially shorter representations while maintaining high reconstruction fidelity and supporting competitive generative modeling. A common approach introduces $K$ learned tokens, also called register tokens, concatenates them with the $P$ image-patch embeddings, and jointly processes the resulting $P+K$ tokens in the encoder. The updated $K$ register tokens are retained as the image representation and passed to the decoder, while the patch tokens are discarded. Although effective, this design increases the model's sequence length from $P$ to $P+K$, adding computational overhead.

Flexible-length tokenizers extend this idea \cite{yamaguchi2025, bachmann2025} by enabling a single model to operate under different token budgets. A commonly used approach is nested dropout \cite{rippel2014}, in which the latent sequence is randomly truncated during training. The decoder must therefore reconstruct images from sequences of varying lengths, allowing the tokenizer to support multiple representation sizes at inference time. 

An alternative design, which we refer to as patch-derived tokens, avoids register tokens altogether and derives its latent tokens from spatially arranged encoder features.
These range from directly retaining or transforming patch tokens to initializing a shorter query sequence through spatial pooling of the encoder feature map \cite{yan2025,fu2026, ye2026}. Nested dropout \cite{rippel2014} can be applied to the resulting encoder sequence to obtain ordered representations of varying lengths. 

Although both bottlenecks have been analyzed in prior work, they have been evaluated under different model architectures, training objectives, quantization schemes, token budgets, and downstream generative models. These differences make it difficult to determine which observed effects can be attributed specifically to the bottleneck design.
We address this gap by isolating the bottleneck design and systematically evaluating its effects on reconstruction quality, generation quality, computational cost, and the spatial organization of the resulting representations. Specifically, our contributions are as follows:
\begin{itemize}
\item We present, to our knowledge, the first controlled comparison of patch-derived and register bottlenecks for flexible-length image tokenization utilizing the same backbone, quantizer, objective and training regime. This allows us to isolate a design choice that is otherwise confounded by differences in model architecture and training setup.

\item We characterize the trade-offs among reconstruction quality, generation quality, and computational cost across a range of token budgets.

\item We analyze the spatial influence of individual latent tokens and
identify consistent spatial organization in both designs. Learned token representations include a subset of spatially diffuse tokens, whereas patch-derived representations exhibit substantially stronger spatial locality.

\end{itemize}

\section{Related Work}

\textbf{Vector-quantized autoencoders}
Vector-quantized autoencoders~\cite{oord2017, oord2019, esser2021} learn discrete compressed representations of images by mapping continuous encoder outputs to entries in a learned codebook. The resulting discrete codes are then passed to the decoder for reconstruction. These models provide an effective framework for learning compact image representations and have become a common basis for generative image models.

\noindent \textbf{Register tokens}
Recent work has explored representing images using compact sets of latent tokens rather than spatially arranged feature maps. TiTok~\cite{yu2024} represents images using a fixed number of latent tokens, which serve as a memory bank that the image tokens of the decoder can retrieve information from via attention. This design decouples the size of the latent representation from the spatial resolution of the input image. One-D-Piece \cite{yamaguchi2025} builds on this idea by introducing an ordered and flexible-length one-dimensional latent representation. This is achieved using tail-token dropout, which can be interpreted as a variant of nested dropout. By dropping suffixes of the latent sequence during training, the model is encouraged to arrange information according to token importance, enabling reconstruction from variable-length prefixes. They further propose a token-contribution analysis which we apply to a controlled comparison of patch-derived and learned-register bottlenecks. FlexTok \cite{bachmann2025} uses learned register tokens with nested dropout and causal attention to produce ordered discrete sequences of variable length. It reconstructs images using a rectified-flow decoder and demonstrates that the resulting tokens support downstream autoregressive generation. SEMANTICIST \cite{wen2025} similarly resamples spatial image features into a sequence of learned concept tokens, but imposes a PCA-like causal ordering and reconstructs images using a diffusion-based decoder. Unlike the discrete representations considered in our study, SEMANTICIST produces continuous latent tokens. Both approaches demonstrate the effectiveness of learned latent queries, but their generative decoders and training objectives differ substantially from deterministic vector-quantized autoencoders.

\noindent \textbf{Patch-derived tokens} ElasticTok \cite{yan2025} instead derives its latent representation directly from spatially arranged encoder tokens and applies nested dropout to support reconstruction from variable-length prefixes. ReTok~\cite{fu2026} uses a related but more elaborate spatially initialized bottleneck: its query tokens are initialized from the encoder's two-dimensional feature map through multi-level average
pooling and are subsequently refined by a Q-Former \cite{carion2020, li2023}.

Existing methods therefore differ not only in how variable-length representations are trained, but also in how their latent tokens are initialized, how they interact with spatial encoder features, which quantization schemes and decoders they use, and how semantic alignment is imposed. These differences make it difficult to isolate the effect
of the bottleneck construction itself. Our work addresses this gap by comparing patch-derived and independently learned register tokens under a shared encoder-decoder architecture, quantizer, nested-dropout schedule, training objective, and downstream autoregressive modeling setup.

\section{Method}

We train autoencoders with variable-length, ordered one-dimensional latent representations. The encoder maps an input image to a sequence of latent tokens, which is discretized using vector quantization. To induce an ordering over this sequence, we apply nested dropout to the latent tokens during training. The autoencoder architecture itself is based on a standard Vision Transformer encoder-decoder backbone.

\subsection{Information Flow}
\begin{figure}[!t]
    \centering
    \includegraphics[width=\textwidth,    height=0.35\textheight,
        keepaspectratio]{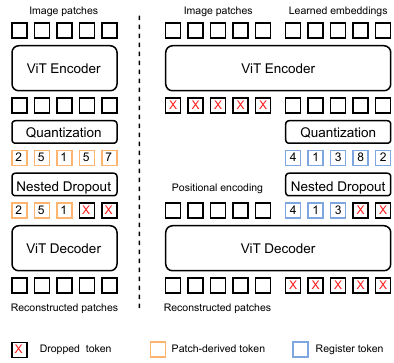}

    \caption{Overview of the investigated architectural designs, with patch-derived tokens on the left and register tokens on the right. Both architectures utilize nested dropout to obtain variable-length image representations. Patch-derived tokens are obtained by transforming image patches into the token representations, whereas register tokens are obtained by transforming dedicated learnable embeddings.} 
    \label{fig:model_architecture}

\end{figure}

For autoencoders with adaptive-length 1D latent representations, two main design choices are used to control the information flow between the encoder and decoder: Patch-derived tokens and register tokens. An overview of both configurations is shown in Figure \ref{fig:model_architecture}.

\noindent \textbf{Patch-derived tokens}
In this configuration, the truncated representation of the encoder, consisting of $r$ tokens, is copied directly to the decoder, initializing the first $r$ decoder tokens. The decoder must learn to transform these tokens into the corresponding image patches. For the remaining zero-initialized tokens, the decoder must infer information by attending to the first $r$ tokens before mapping them to RGB patch representations. Note that the underlying 2D grid structure imposed by rearranging the image patches into a sequence is disrupted only through nested dropout.

\noindent \textbf{Register tokens}
In this approach, the encoder processes the embedded image patches alongside a set of learned register tokens acting as a memory bank that the decoder can attend to. As only the register tokens are passed to the decoder, the encoder is encouraged to store relevant image information in them. As in patch-derived tokens, nested dropout is used to facilitate variable-length decoding. The decoder uses positional embeddings as query tokens that attend to the register tokens to incorporate information about the image. These embeddings are then transformed into RGB patch representations. Because the register tokens are learned parameters rather than spatially arranged image tokens, they do not impose the same inherent 2D grid structure as patch-derived tokens.

\subsection{Training}

\noindent \textbf{Training Objective}
We train the autoencoder following the setup of \cite{xiong2025} optimizing
\begin{equation*}
    \mathcal{L}_{\text{total}} = \mathcal{L}_{\text{VQGAN}} + \lambda \mathcal{L}_{\text{REPA}}.
\end{equation*}
Here, $\mathcal{L}_{\mathrm{VQGAN}}$ combines the $\ell_2$ reconstruction loss computed over image pixels $\mathcal{L}_{\mathrm{recon}}$, a perceptual loss $\mathcal{L}_{\mathrm{perc}}$ \cite{zhang2018, weber2024}, an adversarial loss $\mathcal{L}_{\mathrm{GAN}}$ \cite{esser2021} utilizing a PatchGAN \cite{isola2018} discriminator  and the VQ codebook loss $\mathcal{L}_{\mathrm{VQ}}$ \cite{yu2022}. The final loss term $\mathcal{L}_{\text{REPA}}$ \cite{yu2024b} aligns the intermediate features of a decoder layer with DINOv2-L \cite{oquab2023} features. If we set $\lambda = 1$ we refer to the trained tokenizers as \emph{with REPA}; if $\lambda = 0$ we refer to them as \emph{without REPA}.

\noindent \textbf{Ordered representation} During training we use nested dropout \cite{rippel2014} to encourage the autoencoder to organize information into an ordered sequence. Given an encoder output
$\mathbf{T}=(\mathbf{t}_1,\ldots,\mathbf{t}_n)$, we uniformly sample a retained length $ r \sim \mathcal{U}\{1,\ldots,n\} $
and pass only the first $r$ tokens
$\mathbf{T}_{1:r}=(\mathbf{t}_1,\ldots,\mathbf{t}_r)$ to the decoder. The
remaining tokens $\mathbf{t}_{r+1},\ldots,\mathbf{t}_n$ are discarded. Prior methods restrict $r$ to grid points \cite{fu2026} or sample from an exponentially spaced set \cite{bachmann2025}, which can improve reconstruction performance. In contrast, we sample uniformly over all possible retained lengths, allowing us to evaluate the contribution of each additional latent token independently. Restricting training to selected lengths would make this analysis ambiguous, as multiple tokens would always be introduced jointly.

\section{Experiments}

Our experiments aim at understanding potential performance gaps between Patch-derived tokens and register tokens. Within this section we report our findings regarding their individual reconstruction performance across different token budgets, the performance of downstream generative models and we conduct an analysis of their latent spaces. We show that under comparable training dynamics both approaches yield similar performance regarding reconstruction and downstream generative models. Additionally, we show that the latent representation of both approaches recovers a 2D grid structure. Model details and training specifications can be found in Table \ref{tab:model-architecture} and Table \ref{tab:training-optimization}.
\begin{table*}[t]
\centering
\caption{Inference-speed comparison on \(256 \times 256\) images with a batch size of 128 on a single NVIDIA A100 GPU. The patch-derived token encoder and decoder each process 256 tokens in total, independent of the token budget, as image patches get transformed into tokens and vice versa. The register token encoder processes 256 image-patch embeddings and 256 latent tokens, causing a throughput gap to patch-derived tokens. The register token decoder processes between 257 and 512 tokens in total. Throughput is reported in images per second.}
\label{tab:throughput}

\small
\setlength{\tabcolsep}{6pt}
\renewcommand{\arraystretch}{1.12}

\begin{tabular}{
    @{}
    l
    S[table-format=4.2]
    *{5}{S[table-format=4.2]}
    @{}
}
\toprule
\multirow{2}{*}{\textbf{Token type}} &
{\multirow{2}{*}{\shortstack{\textbf{Encoder}\\
                               \textbf{throughput} }}} &
\multicolumn{5}{c}{\textbf{Decoder throughput }} \\
\cmidrule(lr){3-7}
& &
{\textbf{1}} &
{\textbf{32}} &
{\textbf{64}} &
{\textbf{128}} &
{\textbf{256}} \\
\midrule

Patch-derived
& 1622.9
& \multicolumn{5}{c}{1756.8 (constant)} \\

Register
& 901.3
& 1750.9
& 1573.9
& 1452.6
& 1217.8
& 932.7 \\

\bottomrule
\end{tabular}
\end{table*}

\subsection{Inference Speed}

We measure inference throughput on $256 \times 256$ images using a single
NVIDIA A100 GPU and a batch size of 128. Results are reported in
Table~\ref{tab:throughput}. Although both architectures support
variable-length ordered representations, their computational costs differ
substantially.

For patch-derived tokens, image patches are directly mapped to latent tokens
and back to patch representations. Consequently, both the encoder and decoder
process a fixed sequence of 256 tokens, independent of the retained token
budget, resulting in constant throughput. In contrast, the register token
encoder jointly processes 256 image-patch embeddings and 256 learned register
tokens yielding a substantially lower throughput. The register token decoder processes the retained register tokens together with 256 output queries, such that its sequence length ranges between 257 and 512 tokens. As a result, its
throughput is comparable to that of the patch-derived token decoder at small
token budgets, but decreases substantially as the token budget grows.

\subsection{Reconstruction Performance}

We first compare the reconstruction quality of patch-derived tokens and register tokens across different token budgets. Table \ref{tab:rfid_comparison} reports reconstruction FID (rFID) \cite{heusel2017}, Learned Perceptual Image Patch Similarity (LPIPS) \cite{zhang2018}, mean absolute error (MAE), peak signal-to-noise ratio (PSNR) and structural similarity index measure (SSIM) on the validation split of ImageNet-1K. Without REPA \cite{yu2024b} both models achieve comparable reconstruction performance at high token counts, whereas register tokens seem superior at medium token counts. Surprisingly, patch-derived tokens obtain superior results at low token counts. At very low token counts, however, both approaches fail.
Adding REPA \cite{yu2024b} has mixed results, most notably it deteriorates rFID and LPIPS of register tokens at high token budgets.

\begin{table*}[t]
\centering

\caption{
Reconstruction metrics for the investigated variable-length tokenizers. REPA indicates whether representation alignment was applied during
tokenizer training. All metrics are evaluated on the ImageNet $256 \times 256$ validation split.}

\label{tab:rfid_comparison}
\small
\setlength{\tabcolsep}{4pt}
\renewcommand{\arraystretch}{1.12}

\resizebox{\textwidth}{!}{%
\begin{tabular}{
@{}
l
c
l
*{9}{c}
@{}
}

\toprule

\multirow{2}{*}{\textbf{Token type}} &
\multirow{2}{*}{\textbf{REPA}} &
\multirow{2}{*}{\textbf{Metric}} &
\multicolumn{9}{c}{\textbf{Token count}} \\

\cmidrule(lr){4-12}

& & &
\textbf{1} &
\textbf{4} &
\textbf{8} &
\textbf{16} &
\textbf{32} &
\textbf{64} &
\textbf{128} &
\textbf{192} &
\textbf{256} \\

\midrule

\multirow{10}{*}{Register}

& \multirow{5}{*}{No}
& rFID $\downarrow$
& 344.65
& 223.42
& 77.93
& 7.83
& 3.73
& 2.55
& 1.79
& 1.46
& 1.29 \\

&
& LPIPS $\downarrow$
& 0.600
& 0.560
& 0.518
& 0.455
& 0.395
& 0.341
& 0.293
& 0.271
& 0.258 \\

&
& MAE $\downarrow$
& 0.230 & 0.185 & 0.155 & 0.129 & 0.112 & 0.097 & 0.084 & 0.077 & 0.074 \\

&
& PSNR $\uparrow$
& 11.05 & 12.55 & 13.91 & 15.19 & 16.14 & 17.22 & 18.36 & 19.00 & 19.35 \\

&
& SSIM $\uparrow$
& 0.161 & 0.226 & 0.284 & 0.340 & 0.381 & 0.431 & 0.487 & 0.513 & 0.529 \\

\cmidrule(lr){2-12}

& \multirow{5}{*}{Yes}
& rFID $\downarrow$

& 410.81 & 212.21 & 83.92 & 6.59 & 4.48 & 3.02 & 1.88 & 1.69 & 1.67 \\

&
& LPIPS $\downarrow$
& 0.656 & 0.598 & 0.544 & 0.478 & 0.428 & 0.376 & 0.322 & 0.309 & 0.308 \\

&
& MAE $\downarrow$
& 0.268 & 0.208 & 0.177 & 0.147 & 0.127 & 0.108 & 0.090 & 0.085 & 0.085 \\

&
& PSNR $\uparrow$
& 9.70 & 11.51 & 12.86 & 14.18 & 15.19 & 16.42 & 17.75 & 18.15 & 18.15 \\

&
& SSIM $\uparrow$
& 0.152 & 0.228 & 0.272 & 0.316 & 0.355 & 0.399 & 0.452 & 0.467 & 0.467 \\

\midrule

\multirow{10}{*}{Patch-derived}

& \multirow{5}{*}{No}
& rFID $\downarrow$
& 395.80
& 238.72
& 40.90
& 6.71
& 4.15
& 2.78
& 1.70
& 1.38
& 1.29 \\

&
& LPIPS $\downarrow$
& 0.604 & 0.558 & 0.516 & 0.455 & 0.401 & 0.349 & 0.294 & 0.265 & 0.253 \\

&
& MAE $\downarrow$
& 0.234 & 0.180 & 0.154 & 0.130 & 0.112 & 0.098 & 0.083 & 0.076 & 0.073 \\

&
& PSNR $\uparrow$
& 10.56 & 12.72 & 13.92 & 15.10 & 16.13 & 17.15 & 18.49 & 19.25 & 19.53 \\

&
& SSIM $\uparrow$
& 0.184 & 0.245 & 0.288 & 0.336 & 0.379 & 0.429 & 0.491 & 0.528 & 0.543 \\

\cmidrule(lr){2-12}

& \multirow{5}{*}{Yes}
& rFID $\downarrow$
& 408.10
& 73.89
& 11.60
& 6.18
& 4.80
& 3.44
& 1.92
& 1.33
& 1.21 \\

&
& LPIPS $\downarrow$
& 0.670 & 0.586 & 0.537 & 0.473 & 0.421 & 0.373 & 0.321 & 0.285 & 0.266 \\

&
& MAE $\downarrow$
& 0.273 & 0.205 & 0.176 & 0.145 & 0.123 & 0.107 & 0.091 & 0.082 & 0.077 \\

&
& PSNR $\uparrow$
& 9.61 & 11.85 & 13.00 & 14.31 & 15.43 & 16.39 & 17.62 & 18.59 & 19.11 \\

&
& SSIM $\uparrow$
& 0.163 & 0.236 & 0.274 & 0.319 & 0.360 & 0.404 & 0.458 & 0.500 & 0.523 \\

\bottomrule

\end{tabular}%
}

\end{table*}

\subsection{Generative Performance}

To assess the downstream generative potential of both tokenization approaches, we train class-conditional autoregressive LlamaGen models \cite{sun2024} on the learned latent representations, where we replaced 2D RoPE \cite{lu2024, fang2024} with RoPE \cite{su2024}, as we do not assume a 2D organization of the latent space due to nested dropout. Each model was trained for $1{,}500{,}000$ steps using a batch size of $256$. We report the generation FID (gFID) of $50{,}000$ generated samples relative to the whole training split of ImageNet-1K, following common practice \cite{dhariwal2021}. We report generative performance for token budgets of 16 and above, as reconstruction quality deteriorates substantially at smaller budgets, making these settings less meaningful for downstream generation. The results are presented in Table~\ref{tab:gFID_comparison}.

Increasing the token budget improves performance for most configurations, with the largest gains occurring up to 128 tokens and diminishing returns thereafter. Adding REPA \cite{yu2024b} to the tokenizer training can benefit generative performance when paired with a sufficiently large generative model. The gains are particularly pronounced for patch-derived tokens, which outperform register tokens at small token budgets and achieve comparable performance at larger budgets.

The only clear failure case is LlamaGen-B with register tokens and REPA, whose performance degrades as the token budget increases. We hypothesize that adding REPA alters the structure of the latent space which cannot be utilized properly by a generative model with insufficient capacity.

\begin{table*}[t]
\centering
\caption{
Class-conditional gFID ($\downarrow$) on ImageNet-1K for LlamaGen
models trained with variable-length tokenizers.
REPA indicates whether representation alignment was applied during
tokenizer training. Lower values are better.
}
\label{tab:gfid_comparison}

\small
\setlength{\tabcolsep}{4.5pt}
\renewcommand{\arraystretch}{1.12}

\begin{tabular}{
    @{}
    l
    l
    c
    *{6}{S[table-format=2.2]}
    @{}
}
\toprule
\multirow{2}{*}{\textbf{Token type}} &
\multirow{2}{*}{\textbf{Model}} &
\multirow{2}{*}{\textbf{REPA}} &
\multicolumn{6}{c}{\textbf{Token count}} \\
\cmidrule(lr){4-9}
& & &
{\textbf{16}} &
{\textbf{32}} &
{\textbf{64}} &
{\textbf{128}} &
{\textbf{192}} &
{\textbf{256}} \\
\midrule

\multirow{4}{*}{Register}
& LlamaGen-B & No  & 10.10 & 6.31 & 5.34 & 5.04 & 4.95 & 4.95 \\
& LlamaGen-B & Yes &  9.65 & 8.90 & 9.94 & 13.70 & 15.20 & 15.30 \\
& LlamaGen-L & No  &  9.02 & 5.38 & 4.32 & 3.83 & 3.64 & 3.61 \\
& LlamaGen-L & Yes &  7.22 & 5.34 & 4.08 & 3.22 & 3.11 & 3.11 \\
\midrule

\multirow{4}{*}{Patch-derived}
& LlamaGen-B & No  & 9.90 & 7.85 & 7.03 & 6.46 & 6.39 & 6.38 \\
& LlamaGen-B & Yes & 6.77 & 5.23 & 4.27 & 3.85 & 4.02 & 4.08 \\
& LlamaGen-L & No  & 9.00 & 6.65 & 5.45 & 4.73 & 4.53 & 4.46 \\
& LlamaGen-L & Yes & 6.37 & 4.76 & 3.79 & 3.19 & 3.07 & 3.10 \\
\bottomrule
\end{tabular}
\label{tab:gFID_comparison}
\end{table*}

\subsection{Latent Space Analysis}

\begin{figure*}[t]
\centering
\setlength{\tabcolsep}{4pt}
\renewcommand{\arraystretch}{1.1}

\begin{tabular}{
>{\centering\arraybackslash}m{0.23\textwidth}
>{\centering\arraybackslash}m{0.23\textwidth}
>{\centering\arraybackslash}m{0.23\textwidth}
>{\centering\arraybackslash}m{0.23\textwidth}
}
\multicolumn{2}{c}{\textbf{Patch-derived tokens}}
&
\multicolumn{2}{c}{\textbf{Register tokens}}
\\[1mm]

\shortstack{
    {\small\textbf{Without REPA}}\\[1.5pt]
    {\scriptsize Corrupt token 224}
}
&
\shortstack{
    {\small\textbf{With REPA}}\\[1.5pt]
    {\scriptsize Corrupt token 224}
}
&
\shortstack{
    {\small\textbf{Without REPA}}\\[1.5pt]
    {\scriptsize Corrupt token 142}
}
&
\shortstack{
    {\small\textbf{With REPA}}\\[1.5pt]
    {\scriptsize Corrupt token 145}
}
\\[1mm]

{\tiny
\makebox[0.333\linewidth][c]{Orig.}%
\makebox[0.333\linewidth][c]{Corrupt.}%
\makebox[0.333\linewidth][c]{Diff.}%
}
&
{\tiny
\makebox[0.333\linewidth][c]{Orig.}%
\makebox[0.333\linewidth][c]{Corrupt.}%
\makebox[0.333\linewidth][c]{Diff.}%
}
&
{\tiny
\makebox[0.333\linewidth][c]{Orig.}%
\makebox[0.333\linewidth][c]{Corrupt.}%
\makebox[0.333\linewidth][c]{Diff.}%
}
&
{\tiny
\makebox[0.333\linewidth][c]{Orig.}%
\makebox[0.333\linewidth][c]{Corrupt.}%
\makebox[0.333\linewidth][c]{Diff.}%
}
\\[-1mm]

\includegraphics[
    width=\linewidth
]{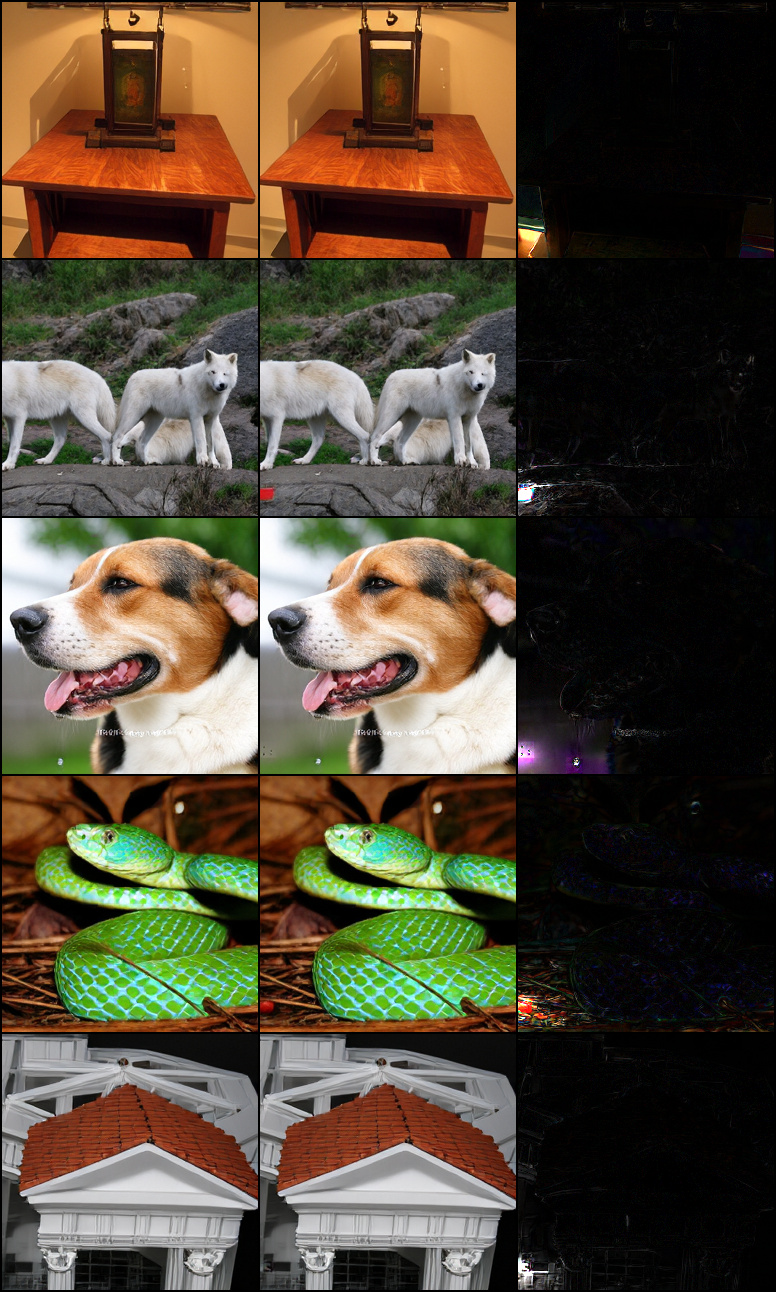}
&
\includegraphics[
    width=\linewidth
]{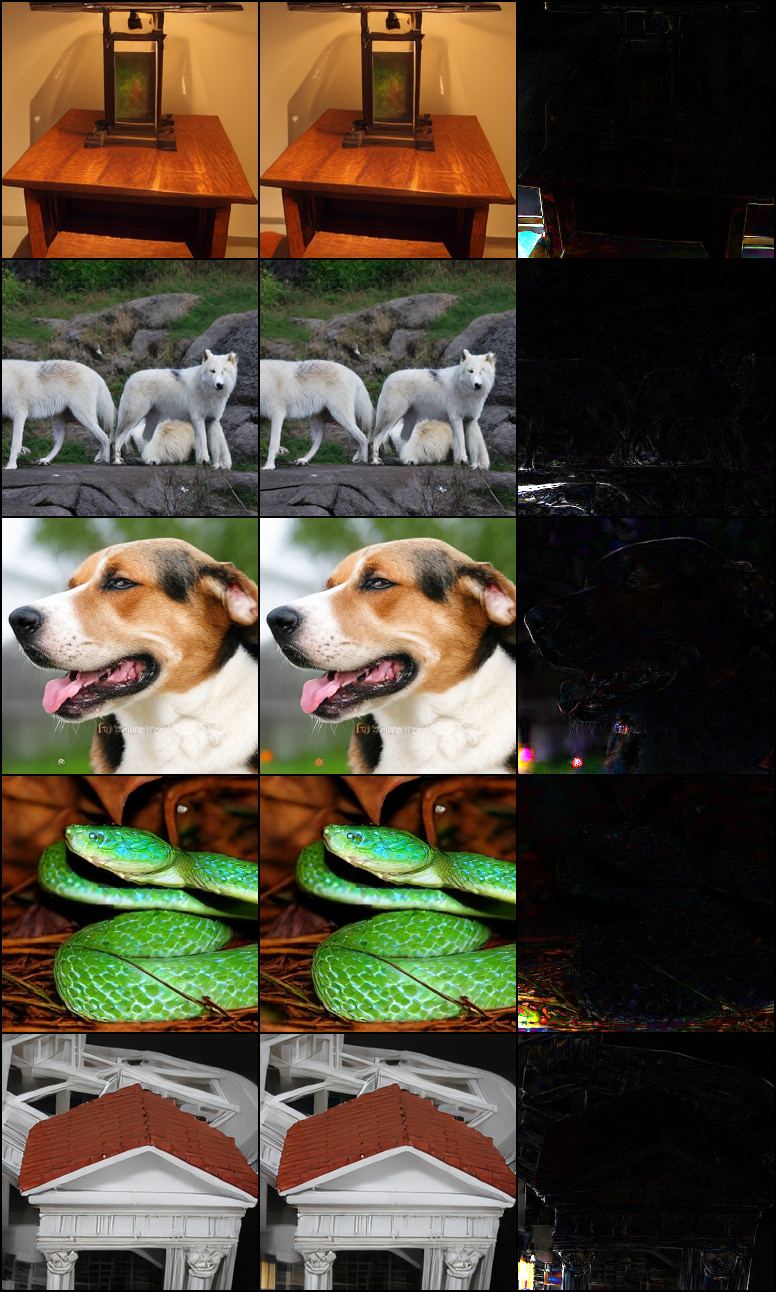}
&
\includegraphics[
    width=\linewidth
]{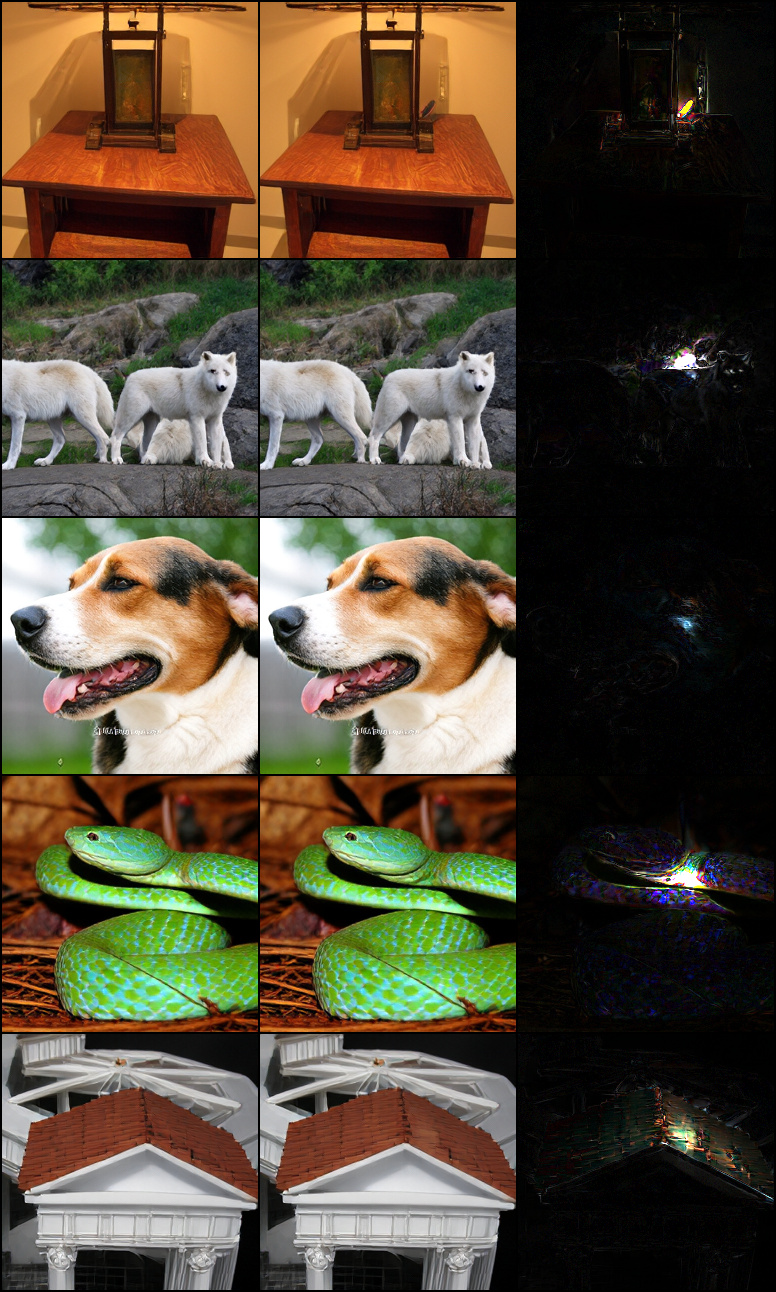}
&
\includegraphics[
    width=\linewidth
]{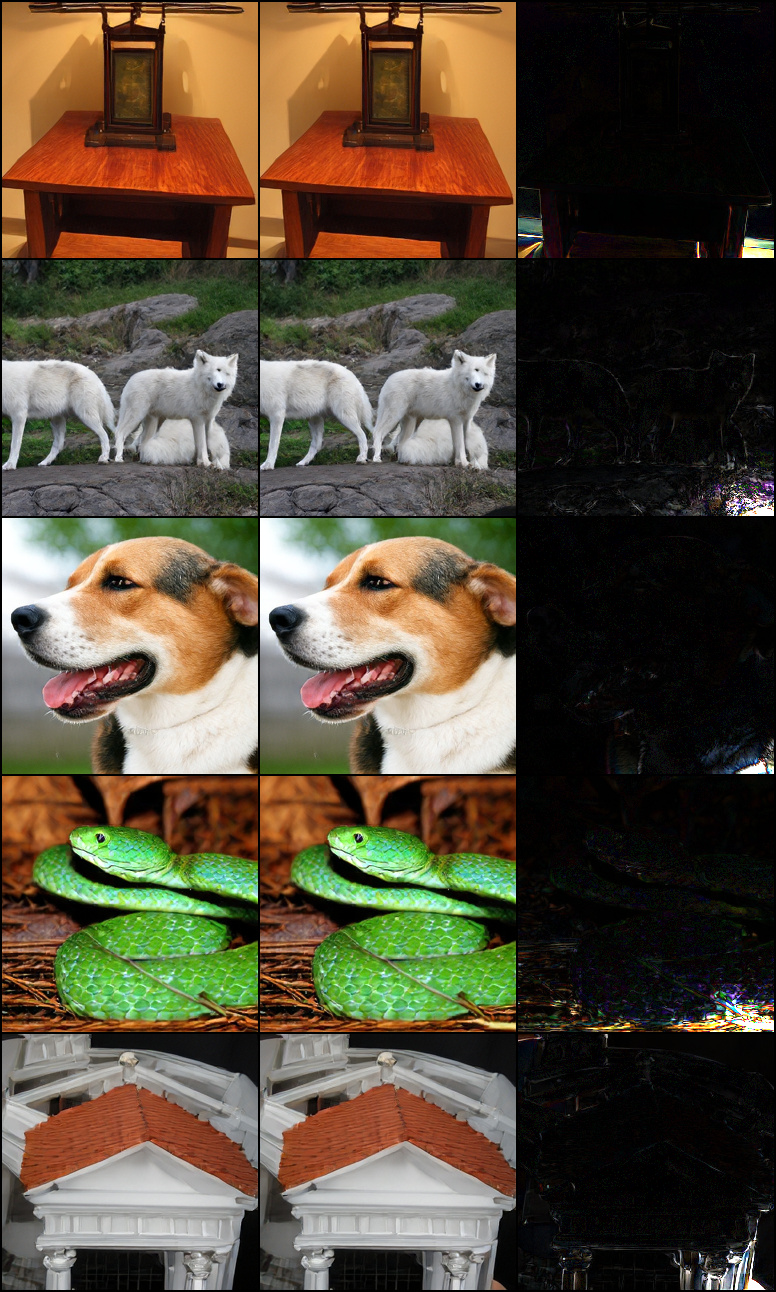}
\\

\end{tabular}

\caption{
Qualitative effect of corrupting a single token.
Within each block, columns show the original reconstruction, the
reconstruction after corrupting the indicated token, and their pixel-wise
difference. For both patch-derived and register tokens, the corruption
predominantly causes spatially localized changes. Although REPA reduces
the strength of this locality, local effects remain visible.
}
\label{fig:corruption-examples}

\end{figure*}

To characterize the spatial influence of individual latent tokens, we
perform a token-corruption analysis. We replace one token at a time with
a randomly sampled codebook entry and measure the squared per-pixel
difference between the corrupted reconstruction and the clean
reconstruction. The resulting sensitivity maps indicate which output
regions are most affected by perturbing each token. Qualitative examples
are shown in Figure~\ref{fig:corruption-examples}.

\begin{figure*}[!p]
\centering

\setlength{\tabcolsep}{1.5pt}
\renewcommand{\arraystretch}{1.0}

\begin{tabular}{
    >{\centering\arraybackslash}m{0.235\textwidth}
    >{\centering\arraybackslash}m{0.235\textwidth}
    >{\centering\arraybackslash}m{0.235\textwidth}
    >{\centering\arraybackslash}m{0.235\textwidth}
}

\multicolumn{2}{c}{\textbf{Patch-derived tokens}}
&
\multicolumn{2}{c}{\textbf{Register tokens}}
\\[0.5mm]

{\small\textbf{Without REPA}}
&
{\small\textbf{With REPA}}
&
{\small\textbf{Without REPA}}
&
{\small\textbf{With REPA}}
\\[0.5mm]

\includegraphics[width=\linewidth]
{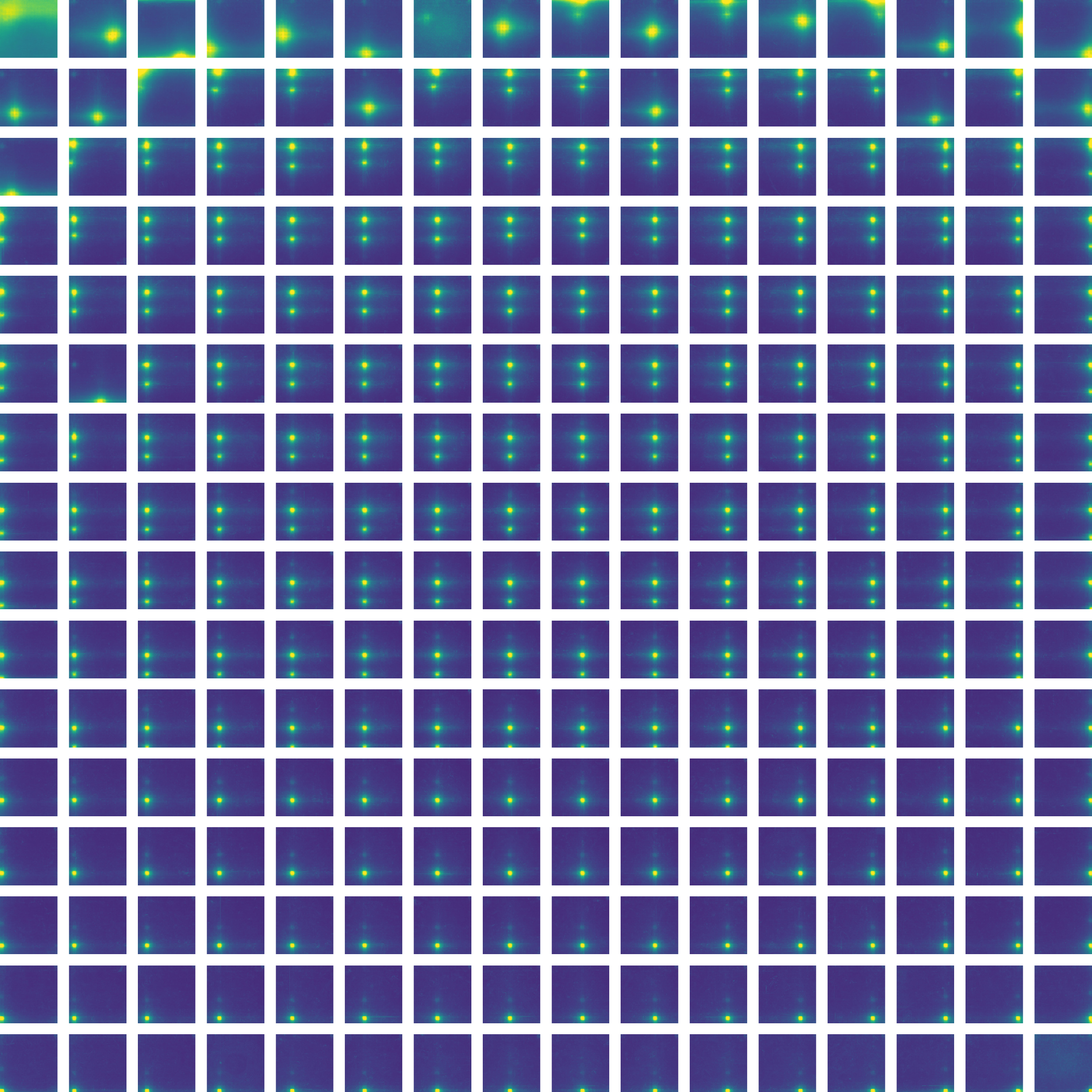}
&
\includegraphics[width=\linewidth]
{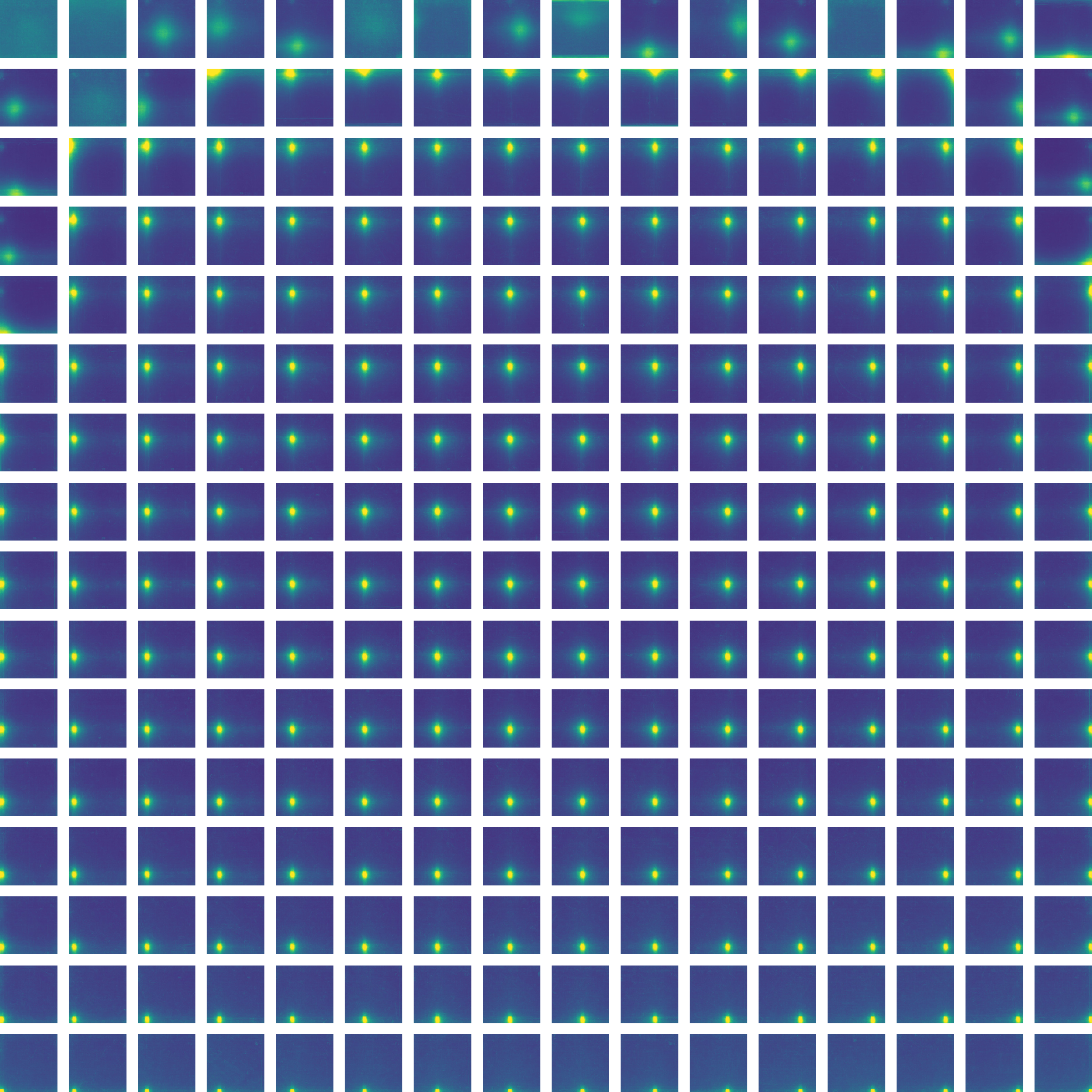}
&
\includegraphics[width=\linewidth]
{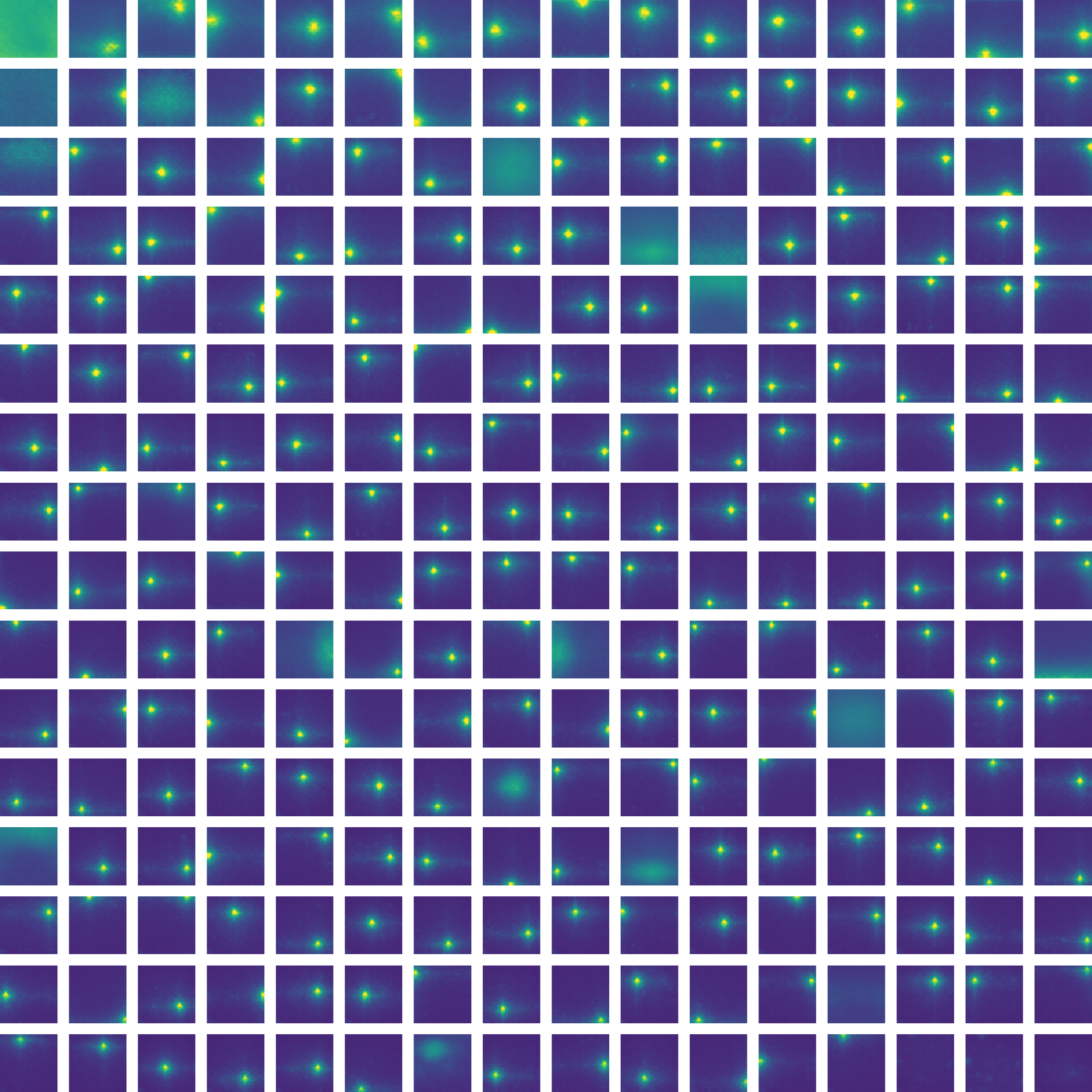}
&
\includegraphics[width=\linewidth]
{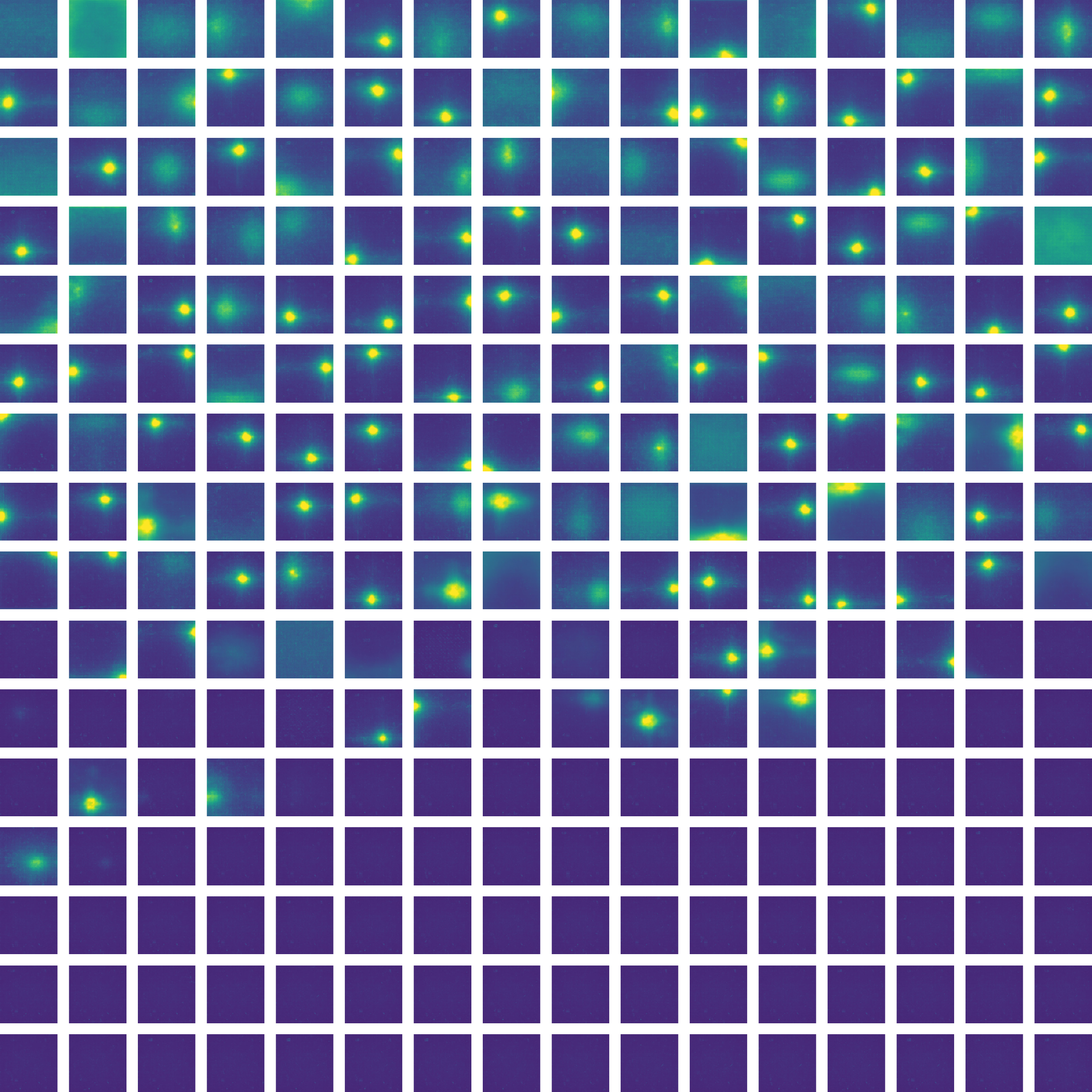}

\end{tabular}

\vspace{-1mm}

\includegraphics[
    width=0.9\textwidth
]{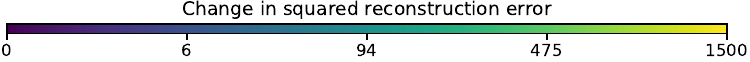}

\vspace{-2mm}

\caption{
Spatial influence of patch-derived tokens and register tokens, trained with and without REPA. Each heatmap shows the pixel-wise change in squared reconstruction error when one token is replaced by a randomly sampled codebook entry. Within each panel, heatmaps are arranged by token index in raster-scan order from top left to bottom right. Without REPA, both tokenization schemes exhibit predominantly localized influence, although some register tokens affect broader image regions. With REPA, patch-derived tokens remain largely localized, whereas many register tokens exhibit substantially more spatially diffuse influence. The color scale uses a power-law normalization to improve visibility. Results are averaged over ImageNet-1K validation images.
}
\label{fig:average-heatmaps-repa-comparison}
\vspace{2mm}

\includegraphics[
    width=\textwidth
]{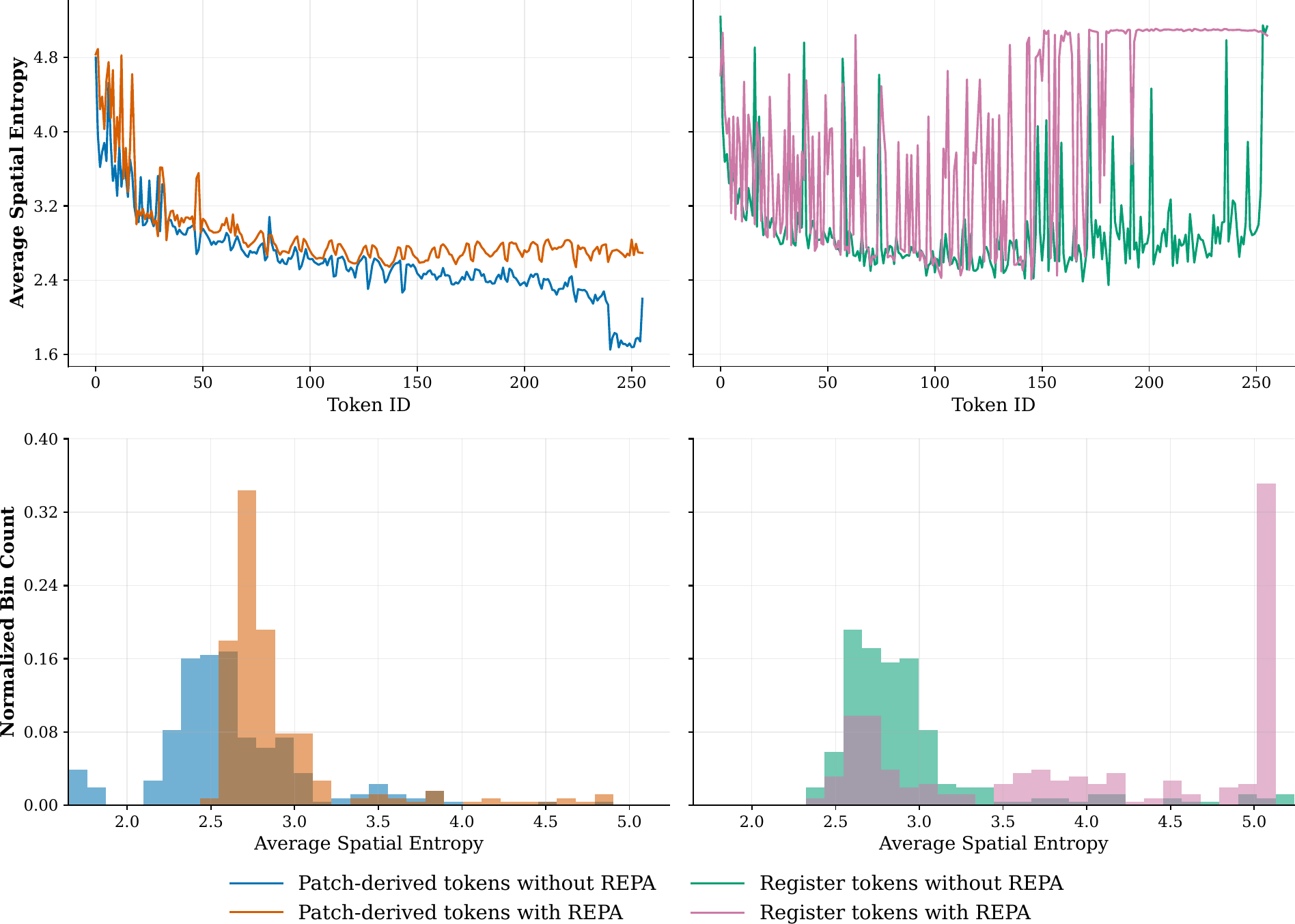}

\vspace{-2mm}

\caption{
Spatial entropy for patch-derived tokens on the left and register tokens on the right. For patch-derived tokens, spatial entropy generally decreases with increasing token ID, indicating progressively stronger spatial concentration; REPA partially mitigates this trend. Register tokens exhibit a different behavior: Their entropy remains comparatively high for later token IDs even without REPA. Adding REPA yields a peak at maximum entropy, which can be attributed to inactive (``dead'') tokens that contribute negligibly to the learned representation.
}
\label{fig:average-entropy-plot}

\end{figure*}

As shown in Figure~\ref{fig:average-heatmaps-repa-comparison}, patch-derived tokens predominantly exhibit localized effects, suggesting that they retain the spatial specialization induced by their initial patch-based organization. Nested dropout partially relaxes this spatial bias, with some of the earliest tokens developing broader spatial influence. Register tokens also frequently develop spatially localized influence, despite having no predefined token-to-patch
correspondence. This indicates  that spatial specialization can emerge given the architecture design and reconstruction objective. 
REPA largely preserves the localized behavior of patch-derived tokens, but leads to more spatially diffuse effects for many register tokens. We also observe a substantial number of register tokens with low total corruption sensitivity; replacing them produces only small changes in the reconstructed image. The prevalence of such low-sensitivity tokens at later sequence positions is consistent with the diminishing improvements in rFID at high token budgets. 

To quantify the spatial concentration of each token's influence, we
aggregate its reconstruction-difference map over the output
patch grid. Let \(s_{i,j}\) denote the sum of the reconstruction-difference values in output patch \(j\) after corrupting token \(i\). We normalize the values as
\[
q_{i,j} =
\frac{s_{i,j}}{\sum_{j'} s_{i,j'}}
\]
and compute the spatial entropy
\[
H_i =
-\sum_{j=1}^{P} q_{i,j}\log q_{i,j},
\]
where \(P\) is the number of output patches. Lower entropy indicates
spatially concentrated influence, while higher entropy indicates more
spatially diffuse influence. Figure \ref{fig:average-entropy-plot} shows that register tokens are generally more spatially diffuse
than patch-derived tokens and that REPA increases this trend.

\begin{table}[ht]
\centering
\small

\begin{minipage}{0.78\textwidth}
\centering
\captionof{table}{Model architecture and quantization configuration.}
\label{tab:model-architecture}

\begin{tabular}{
    @{}
    >{\raggedright\arraybackslash}p{0.58\linewidth}
    >{\raggedright\arraybackslash}p{0.30\linewidth}
    @{}
}
\toprule
Parameter & Value \\
\midrule
Patch size                     & 16 \\
Input image size               & $256 \times 256$ \\
Number of image patches        & $256$ \\
Encoder depth                  & 12 \\
Encoder dimension              & 768 \\
Decoder depth                  & 12 \\
Decoder dimension              & 768 \\
Nested-dropout mode            & Uniform \\
Quantization method            & VQ \cite{oord2017} \\
Codebook size                  & 16{,}384 \\
Codebook vector dimension      & 8 \\
Total parameters               & 195,962,460 \\
\makecell[l]{Number of latent tokens\\
             (only for register tokens)} & 256 \\
Register token parameters      & 589{,}824 \\
\bottomrule
\end{tabular}
\end{minipage}

\vspace{1.5em}

\begin{minipage}{0.78\textwidth}
\centering
\captionof{table}{Training and optimization configuration.}
\label{tab:training-optimization}

\begin{tabular}{
    @{}
    >{\raggedright\arraybackslash}p{0.58\linewidth}
    >{\raggedright\arraybackslash}p{0.30\linewidth}
    @{}
}
\toprule
Parameter & Value \\
\midrule
Training steps               & 650K \\
Dataset                      & ImageNet-1K \cite{deng2009} \\
Image resolution             & $256 \times 256$ \\
Batch size                   & 512 \\
Optimizer                    & AdamW \cite{loshchilov2017} \\
Learning rate                & $10^{-4}$ \\
Optimizer parameters         & \makecell[tl]{
    $\beta_1 = 0.9$,\\
    $\beta_2 = 0.999$
} \\
Weight decay                 & $10^{-4}$ \\
Precision                    & \texttt{bfloat16} \\
MSE loss weight              & $1.0$ \\
Perceptual loss weight       & $1.1$ \\
Discriminator loss weight    & $0.1$ \\
VQ Commitment loss weight    & $0.25$ \\
LeCam \cite{tseng2021} regularization weight  & $10^{-3}$ \\
REPA loss weight             & $0.0$ or $1.0$ \\
REPA layer (decoder)         & 1 \\
REPA model                   & DINOv2-L \\
REPA projection              & 3-layer MLP \\
\bottomrule
\end{tabular}
\end{minipage}

\end{table}

\clearpage
\section{Conclusion}

We investigated two approaches for learning compact visual token representations: Patch-derived tokens, which inherit an explicit spatial organization, and register tokens, which are free to develop a latent structure during training. By combining both approaches with nested dropout, we obtain representations that support variable token budgets and can be used for both reconstruction and downstream autoregressive generation.

Our generative evaluation shows that increasing the token budget improves performance up to approximately 128 -- 192 tokens, after which gains largely saturate. REPA further improves generative quality when the downstream model has sufficient capacity and is particularly effective for patch-derived tokens, which perform strongly even at small token budgets. Register tokens achieve comparable performance with a sufficiently large generative model, but their combination with REPA is less robust when the generative model's capacity is limited.

The token-corruption analysis provides additional insight into the learned latent organization. Patch-derived tokens largely retain spatially localized influence, while register tokens also develop spatial specialization despite having no explicitly imposed two-dimensional layout. REPA can produce tokens with broader spatial influence, especially among register tokens. Overall, our results suggest that both structured and unconstrained tokenization schemes can learn effective compact image representations, while their inductive biases lead to different trade-offs in spatial organization and downstream generative performance.

Although most aspects of the experimental setup are tightly controlled across tokenizer variants, the comparison is not compute-matched. Register tokens require more computation due to the increased number of tokens processed per image.
An explicit compute-matched comparison could disentangle the effects of bottleneck design from differences in training compute and clarify whether the additional compute contributes to the performance gap observed without REPA.

Further, our work focused entirely on autoregressive models for the downstream generative task, as nested dropout encourages the learning of an ordered representation. It remains open whether non-autoregressive models can effectively exploit such an ordering.

\section*{Acknowledgement}

The research leading to these results is funded by the German Federal Ministry for Economic Affairs and Energy within the project ``NXT GEN AI METHODS -- Generative Methoden für Perzeption, Prädiktion und Planung''. The authors would like to thank the consortium for the successful cooperation.

\bibliographystyle{splncs04}
\bibliography{6_references}

\end{document}